\documentclass[10pt,twocolumn,letterpaper]{article}

\usepackage[pagenumbers]{cvpr} % To force page numbers, e.g. for an arXiv version
\usepackage[accsupp]{axessibility}
\usepackage{booktabs}   % 三线表
\usepackage{array}      % 列对齐
\usepackage{caption}    % 精确控制标题
\usepackage{float}      % 用于[H]定位

\usepackage{booktabs}
\definecolor{cvprblue}{rgb}{0.21,0.49,0.74}
\usepackage[pagebackref,breaklinks,colorlinks,allcolors=cvprblue]{hyperref}
\usepackage{caption}

\usepackage[table]{xcolor}
\usepackage{graphicx}
\usepackage{booktabs}
\def\paperID{37263} % *** Enter the Paper ID here
\def\confName{CVPR}
\def\confYear{2026}

\title{Sparsity-Aware Voxel Attention and Foreground Modulation for 3D Semantic Scene Completion}

\author{
Yu Xue\thanks{Equal contribution.} \qquad
Longjun Gao\footnotemark[1] \qquad
Yuanqi Su\thanks{Corresponding author: yuanqisu@mail.xjtu.edu.cn} \qquad
HaoAng Lu \qquad
Xiaoning Zhang\\
Xi’an Jiaotong University, China\\
{Project Page: \href{https://github.com/xyandtyh/VoxSAMNet}{https://github.com/xyandtyh/VoxSAMNet}}
}

\begin{document}
\maketitle
% =========================
% Common table color config
% =========================
\newcommand{\clsbox}[1]{\textcolor{#1}{\rule{1.3ex}{1.3ex}}}

\newcommand{\cls}[2]{%
  \shortstack[c]{%
    \rotatebox{90}{\textbf{#2}}\\[-0.2ex]
    \clsbox{#1}%
  }%
}
% ----- shared colors from your visualization script -----
\definecolor{clsCar}{RGB}{100,150,245}
\definecolor{clsBicycle}{RGB}{100,230,245}
\definecolor{clsMotorcycle}{RGB}{30,60,150}
\definecolor{clsTruck}{RGB}{80,30,180}
\definecolor{clsOtherVeh}{RGB}{100,80,250}
\definecolor{clsPerson}{RGB}{255,30,30}
\definecolor{clsBicyclist}{RGB}{255,40,200}
\definecolor{clsMotorcyclist}{RGB}{150,30,90}
\definecolor{clsRoad}{RGB}{255,0,255}
\definecolor{clsParking}{RGB}{255,150,255}
\definecolor{clsSidewalk}{RGB}{75,0,75}
\definecolor{clsOtherGround}{RGB}{175,0,75}
\definecolor{clsBuilding}{RGB}{255,200,0}
\definecolor{clsFence}{RGB}{255,120,50}
\definecolor{clsVegetation}{RGB}{0,175,0}
\definecolor{clsTrunk}{RGB}{135,60,0}
\definecolor{clsTerrain}{RGB}{150,240,80}
\definecolor{clsPole}{RGB}{255,240,150}
\definecolor{clsTrafficSign}{RGB}{255,0,0}

% ----- KITTI-360 only classes -----
% Since your KITTI-360 visualization branch is coarse-grained (10 classes),
% these two colors are assigned in a consistent style:
% other-struct.: structure-like neutral gray
% other-obj.: other-category magenta-purple tone
\definecolor{clsOtherStruct}{RGB}{0,150,255}
\definecolor{clsOtherObj}{RGB}{255,255,50}

% ----- small color square -----
\providecommand{\clsbox}[1]{%
  \textcolor{#1}{\rule{1.0ex}{1.0ex}}%
}

% ----- rotated header: text + color box -----
\providecommand{\cls}[2]{%
  \rotatebox{90}{\textbf{#2}\,\clsbox{#1}}%
}
\begin{abstract}
Monocular Semantic Scene Completion (SSC) aims to reconstruct complete 3D semantic scenes from a single RGB image, offering a cost-effective solution for autonomous driving and robotics. However, the inherently imbalanced nature of voxel distributions—where over 93\% of voxels are empty and foreground classes are rare—poses significant challenges. Existing methods often suffer from redundant emphasis on uninformative voxels and poor generalization to long-tailed categories.

To address these issues, we propose \textbf{VoxSAMNet} (Voxel Sparsity-Aware Modulation Network), a unified framework that explicitly models voxel sparsity and semantic imbalance. Our approach introduces: (1) a \textbf{Dummy Shortcut for Feature Refinement (DSFR)} module that bypasses empty voxels via a shared dummy node while refining occupied ones with deformable attention; (2) a \textbf{Foreground Modulation Strategy} combining \textit{Foreground Dropout (FD)} and \textit{Text-Guided Image Filter (TGIF)} to alleviate overfitting and enhance class-relevant features.

Extensive experiments on the public benchmarks SemanticKITTI and SSCBench-KITTI-360 demonstrate that VoxSAMNet achieves state-of-the-art performance, surpassing prior monocular and stereo baselines with mIoU scores of 18.2\% and 20.2\%, respectively. Our results highlight the importance of sparsity-aware and semantics-guided design for efficient and accurate 3D scene completion, offering a promising direction for future research.
\end{abstract}    
\section{Introduction}
\begin{figure*}[h!]
    \centering
    %\parbox{6.5in}
    \includegraphics[width=0.85\linewidth]{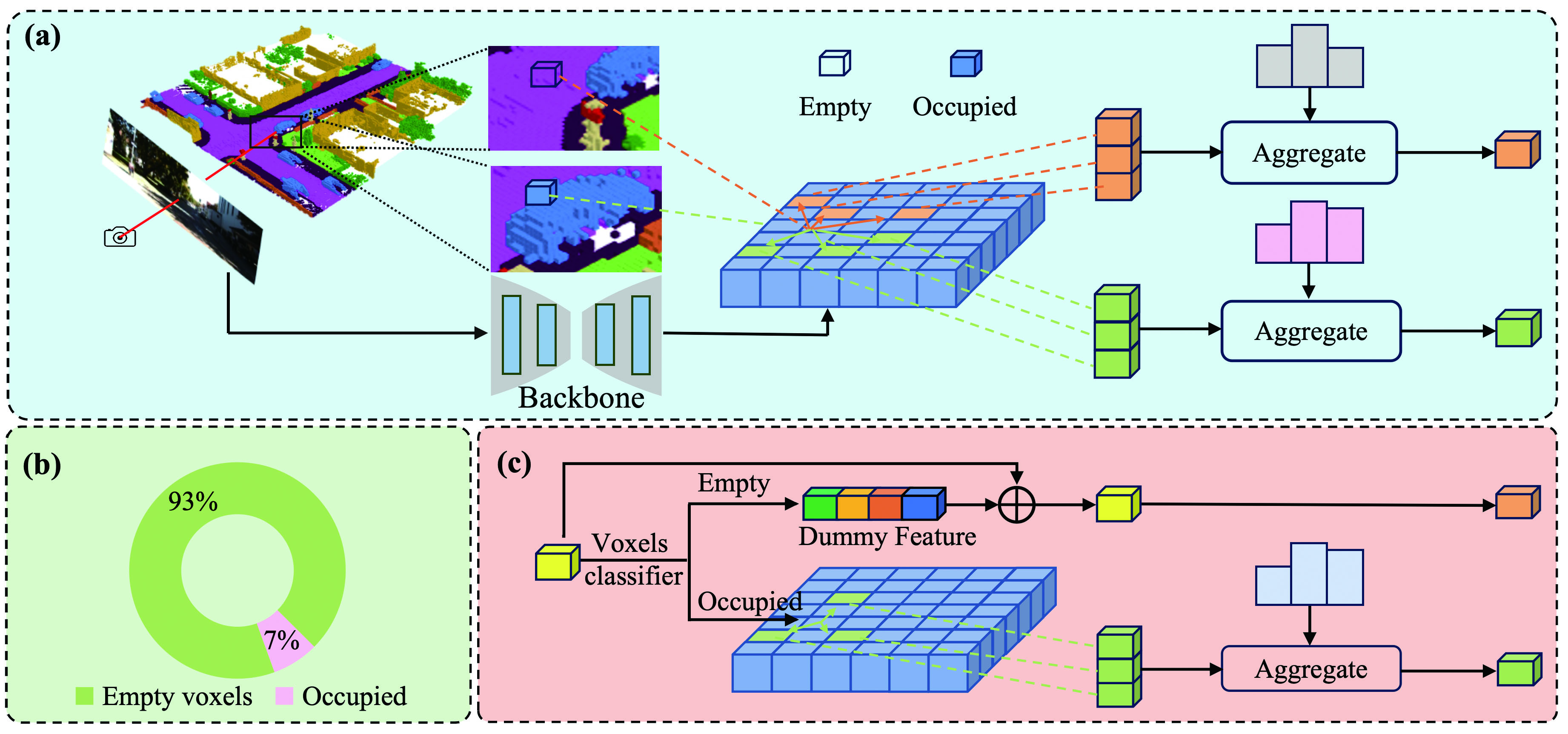}
    \caption{Motivation for VoxSAMNet. (a) Current methods such as BEVFormer~\cite{li2022bevformer} and Et-former~\cite{liang2024former} use deformable attention to update features for both empty and occupied voxels. (b) The ratio of occupied and empty voxels in the SemanticKITTI~\cite{behley2019semantickitti} dataset. (c) We designed a branch that uses two different feature update methods for empty voxels and occupied voxels. }
    \label{fig:motivation}
\end{figure*}
Monocular Semantic Scene Completion (SSC)~\cite{song2017semantic, cao2022monoscene, li2023voxformer, yu2024context, liu2025disentangling,wang2024not,zheng2024monoocc, lin2025orthalign} aims to reconstruct complete 3D semantic scenes from a single RGB image. This task plays a pivotal role in autonomous driving and robotics, offering cost-effective perception without LiDAR or stereo inputs.

However, SSC remains highly challenging due to the \textbf{severely imbalanced voxel distribution} in 3D space. For instance, in the SemanticKITTI dataset~\cite{behley2019semantickitti}, over \textbf{93\%} of voxels are empty, and among the remaining \textbf{7\%} occupied voxels, foreground semantic classes (e.g., pedestrians, cyclists) account for only a small fraction. This dual imbalance—both spatial and semantic—leads to two major issues: (1) excessive emphasis wasted on uninformative empty voxels, and (2) poor generalization caused by overfitting to rare foreground categories.

Despite recent progress, existing camera-based SSC methods, including both monocular and stereo-based approaches, still fall short in addressing these challenges.

\textbf{MonoScene}~\cite{cao2022monoscene}, for example, employs a 3D U-Net with sparse convolutions to reduce computational load. However, due to its uniform voxel processing, it still applies convolutions across the entire space, including the vast majority of empty voxels—resulting in inefficient computation and diluted learning signals for informative regions.

\textbf{BEVFormer}~\cite{li2022bevformer} introduces deformable attention~\cite{zhu2020deformable}, projecting voxels to the 2D image plane and aggregating features via learned offsets. While this enables spatially adaptive sampling, it suffers from two critical limitations: (1) \textbf{feature ambiguity}, as multiple voxels along the same viewing ray are projected to the same 2D location, and (2) \textbf{redundant computation}, since attention is applied indiscriminately to both empty and occupied voxels.

Other approaches, such as \textbf{VoxFormer}~\cite{li2023voxformer}, \textbf{CGFormer}~\cite{yu2024context} and \textbf{Occdepth}~\cite{miao2023occdepth}, attempt to construct voxel queries from depth priors. However, they often rely on fixed sampling grids or global attention, lacking spatial adaptiveness and failing to incorporate explicit voxel occupancy awareness. Importantly, none of these methods treat empty and occupied voxels differently during feature refinement, limiting their ability to concentrate computation on semantically valuable regions.

In summary, current methods suffer from three key limitations:
\vspace{-0.3em}
\begin{list}{\textbullet}{%
  \setlength{\leftmargin}{2em}%
  \setlength{\labelsep}{0.5em}%
  \setlength{\itemsep}{0.2ex}%
  \setlength{\parsep}{0pt}%
  \setlength{\topsep}{0.4ex}%
}
  \item \textbf{Redundant emphasis} on empty voxels due to uniform processing paradigms,
  \item \textbf{Semantic dilution} caused by projection-induced feature ambiguity,
  \item \textbf{Foreground underrepresentation} due to lack of targeted enhancement for rare classes.
\end{list}

To address these shortcomings, we propose \textbf{VoxSAMNet} (Voxel Sparsity-Aware Modulation Network), a unified monocular SSC framework that explicitly models voxel sparsity and foreground semantic imbalance through selective computation and semantic-aware modulation. VoxSAMNet introduces the following three innovations:

\textbf{1) Dummy Shortcut for Feature Refinement (DSFR)}: a sparsity-aware attention module that distinguishes between empty and occupied voxels. Empty voxels are mapped to a shared learnable dummy node, avoiding unnecessary computation, while occupied voxels are refined using multi-head deformable attention. This design enables efficient and focused feature aggregation.

\textbf{2) Foreground Modulation Strategy}: To mitigate overfitting under long-tailed distributions, we introduce two synergistic components. \textit{foreground dropout} randomly suppresses supervision signals for foreground classes during training, enhancing generalization. \textit{Text-Guided Image Filter (TGIF)}, inspired by CLIP~\cite{radford2021learning}, GroundingDINO~\cite{liu2024grounding}, GLIP~\cite{li2022grounded} and LLaVA~\cite{liu2023visual}, uses text prompts to downweight dropped categories in the 2D feature space, reinforcing the remaining class representations.

We validate VoxSAMNet on SemanticKITTI~\cite{behley2019semantickitti} and SSCBench-KITTI-360~\cite{liao2022kitti,li2024sscbench, lin2026hidden}, achieving state-of-the-art performance. Notably, our method outperforms not only existing monocular approaches but also recent stereo and multi-view baselines.

\noindent\textbf{Our main contributions are summarized as follows:}
\begin{itemize}
\item We propose \textbf{VoxSAMNet}, a unified monocular SSC framework that tackles voxel sparsity and semantic imbalance through sparsity-aware attention and semantic-guided modulation.
\item We introduce \textbf{DSFR}, a selective refinement module that bypasses empty voxels via a shared dummy node while enhancing occupied voxels with deformable attention.
\item We design a \textbf{foreground modulation strategy} combining FD and TGIF, which alleviates overfitting on rare classes and enhances semantic consistency.
\end{itemize}

\section{Related Work}
\subsection{Monocular Semantic Scene Completion}
Monocular Semantic Scene Completion (SSC) has made significant progress in reconstructing 3D geometry and semantics from a single RGB image, with broad applications in autonomous driving. MonoScene~\cite{cao2022monoscene} proposes a two-stage pipeline combining 2D and 3D U-Nets~\cite{cciccek20163d}, connected via an optics-inspired 2D-to-3D projection module. It improves spatio-semantic consistency through 3D context priors and introduces novel loss functions such as global scene and frustum losses.

Building on this, NDC-Scene~\cite{yao2023ndc} operates in Normalized Device Coordinates (NDC) space to mitigate projection ambiguity and computational imbalance, progressively restoring depth to lift 2D features into 3D. DepthSSC~\cite{yao2023depthssc} enhances object boundary quality via spatial transformation and geometry-aware voxelization. These methods demonstrate notable advances in 2D-to-3D feature lifting and semantic fusion~\cite{li2023stereoscene,li2023fb,wang2022detr3d,zhang2026coherencetrapmllmcraftednarratives, zhang2026igasa}.

However, most of these approaches apply uniform processing across the entire voxel space, overlooking the extreme sparsity and long-tailed semantic distribution of 3D scenes. Our work addresses this by explicitly modeling voxel sparsity and semantic imbalance during feature refinement.
\vspace{-0.3em}
\subsection{Sparsity-Aware Scene Modeling}
The dominance of empty voxels in 3D grids motivates sparsity-aware design for computational efficiency. VoxFormer~\cite{li2023voxformer} initiates inference from a sparse set of depth-predicted voxel queries and employs masked autoencoding to propagate features to the full voxel space. CGFormer~\cite{yu2024context} extends this by incorporating context and geometry-aware voxel queries and 3D deformable cross-attention to improve depth alignment.

BEVFormer~\cite{li2022bevformer}, though designed for bird’s-eye-view representation in multi-camera systems, introduces deformable attention to focus computation on informative regions, offering transferable insights to SSC. SGN~\cite{mei2024camera} adopts a dense-sparse-dense structure: it selects semantic-aware seed voxels from estimated depth and refines them via hybrid guidance and multi-scale attention.

These methods highlight the benefits of sparse voxel representation and selective feature aggregation. Nonetheless, few methods explicitly differentiate between empty and occupied voxels during refinement. Our method introduces a dummy shortcut mechanism to bypass empty voxels and concentrate computation on informative regions.

\subsection{Text Guided Feature Modulation}
The integration of textual modalities into 3D scene understanding is an emerging research direction. VLScene~\cite{wang2025vlscene} introduces visual-language distillation for SSC, aligning image and text features to supervise semantic completion. CDScene~\cite{wang2025vision} fuses visual and language representations to disentangle dynamic/static components, guided by depth modeling and multi-modal fusion. MMSG~\cite{he2025multi} constructs multi-modal scene graphs with large language models to support embodied navigation.

While these approaches demonstrate the potential of language-guided 3D perception, most operate at the scene or object level rather than voxel-level refinement. Inspired by CLIP~\cite{radford2021learning}, GroundingDINO~\cite{liu2024grounding}, and GLIP~\cite{li2022grounded}, our approach introduces a Text-Guided Image Filter (TGIF) that modulates 2D features based on prompted categories, enhancing robustness under long-tailed distributions.

\section{Methodology}

\subsection{Method Overview}
\begin{figure*}[h!]
    \centering
    %\parbox{6.5in}
    {\includegraphics[width=.9\linewidth]{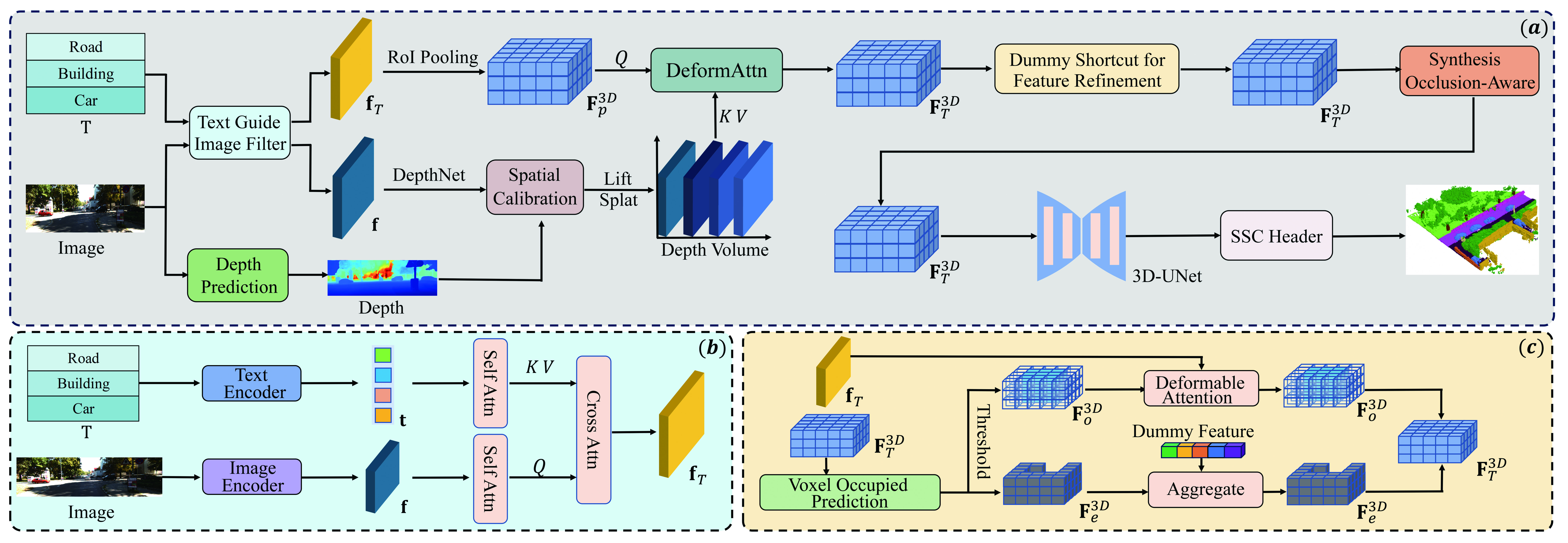}}
    \caption{Overview of VoxSAMNet and the structure of its modules. (a) The pipeline flow of the proposed VoxSAMNet. (b) The specific structure of the Text-Guided Image Filter. (c) The specific structure of Dummy Shortcut for Feature Refinement.}
    \label{fig:architecture}
\end{figure*}

Given an input RGB image, our goal is to reconstruct a complete 3D semantic scene by predicting a voxel-wise semantic occupancy grid. The proposed framework, \textbf{VoxSAMNet} (illustrated in Figure~\ref{fig:architecture}), introduces key innovations to address the fundamental challenges of long-tailed class distribution and the inherent sparsity of 3D spaces. The semantic predictions are supervised by cross-entropy loss with the corresponding labels. Our pipeline operates in three synergistic stages:

\paragraph{Stage I: Text-Guided 3D Initialization (Sec.~\ref{sec:tgif}).}  
To improve semantic discrimination under long-tailed distributions, we introduce a \textbf{Text-Guided Image Filter (TGIF)}. TGIF leverages category-aware language prompts to selectively activate relevant foreground features on the 2D image plane while suppressing residual responses from the dropped categories. These filtered 2D features are subsequently lifted into 3D space guided by dense depth estimation, forming an initial, semantically-focused 3D voxel representation.

\paragraph{Stage II: Dummy Shortcut for Feature Refinement (Sec.~\ref{sec:dsfr}).}  
Recognizing that most of the 3D space is empty, we propose the \textbf{Dummy Shortcut for Feature Refinement (DSFR)} to optimize computational efficiency. A voxel-wise classifier first predicts the occupancy status of the initial 3D volume. While occupied voxels undergo rigorous refinement via 3D deformable attention against the 2D feature map, empty voxels are routed through a lightweight dummy shortcut. This selective mechanism ensures that computational resources are concentrated on informative regions.

\paragraph{Stage III: Completion via Denoising MAE and 3D U-Net.}  
To recover occluded or unseen regions, we adopt a masked voxel modeling strategy based on \textbf{OcclusionMAE}~\cite{Lu_2025_ICCV,he2022masked}. Noise is applied to visible voxel features, and a 3D U-Net is used to infer missing content via contextual reasoning. This stage enhances voxel completeness and global consistency, particularly in sparsely observed areas. The final refined features are then decoded to predict the semantic occupancy and supervised by our joint loss functions (Sec.~\ref{sec:loss}).

\subsection{Text-Guided 3D Initialization} \label{sec:tgif}

To construct semantically meaningful 3D voxel features from 2D image inputs, we propose a text-guided 3D initialization framework that combines visual-textual filtering and voxel-aware feature projection. This component addresses the long-tailed distribution of foreground categories, a primary challenge in SSC. Our approach integrates two synergistic mechanisms: foreground dropout and the \textbf{Text-Guided Image Filter (TGIF)}.

\vspace{0.2em}\noindent\textbf{Text-Guided Image Filter (TGIF).}\hspace{0.4em}
Due to the class imbalance inherent in real-world 3D scenes, semantic segmentation models often overfit to frequent foreground categories (e.g., \emph{car}, \emph{person}). To mitigate this, we introduce foreground dropout as a regularization technique inspired by Dropout. During training, the foreground dropout randomly masks the ground-truth labels of foreground instances with a probability $p$, replacing them with an empty voxel index (e.g., 0). These masked foreground labels are subsequently reassigned to the empty class, thereby encouraging the learning of more generalizable representations.

While foreground dropout reassigns selected foreground labels to the empty class, residual semantic responses of those dropped categories may still remain in the 2D feature space. To address this issue, TGIF is designed to further suppress such irrelevant responses through cross-modal guidance. Specifically, for each training sample, a \emph{category-aware} text prompt $T$ is generated from the foreground categories retained after dropout (e.g., ``road, building, terrain''). Let $\mathbf{f}$ denote the 2D image feature extracted by a vision backbone. This prompt $T$ is encoded by a language encoder and fused with the image features $\mathbf{f}$ through a combination of self-attention and cross-attention (Figure~\ref{fig:architecture}(b)). The TGIF module modulates $\mathbf{f}$ to produce category-aware features $\mathbf{f}_T$:
\begin{equation}
    \mathbf{f}_T = \text{TGIF}(\mathbf{f}, T)
\end{equation}
The resulting filtered feature map $\mathbf{f}_T$ preserves semantics aligned with the retained categories, while suppressing residual interference from the dropped ones, thereby facilitating downstream 3D lifting.

\vspace{0.2em}\noindent\textbf{3D Feature Initialization.} 
Given the filtered 2D feature map $\mathbf{f}_T$, we lift it into a 3D voxel grid to obtain the initial 3D representation. The lifting process begins by projecting each 3D voxel center ${P}=(x, y, z)^\top$ to the image plane using extrinsic and intrinsic camera parameters. The transformation is defined as:
\begin{equation}
    \widetilde{\mathbf{u}} = {K}[{R}|{t}]{P}
\end{equation}
where ${K}$ denotes the camera intrinsic matrix, and $[{R}|{t}]$ represents the extrinsic parameters. The projected 2D coordinates $\widetilde{\mathbf{u}}$ correspond to pixel locations $(u, v)^\top$ in the image.

Next, a binary field-of-view mask $\mathrm{Mask}(u,v)$ is applied to filter out voxels that fall outside the image region. For valid voxels, we extract 2D features through ROI pooling~\cite{girshick2015fast}, and then refine them using 3D deformable attention~\cite{zhu2020deformable,Li_2023_ICCV,wang2022detr3d,zhang2023monodetr, zhang2026cmhanet, zhang2026chain} based on the estimated depth volume $D_p$. The depth volume is formed by lifting dense depth estimated from a DepthNet~\cite{Anunay2021DepthNet} via LSS~\cite{philion2020lift}. The initialized 3D feature $\mathbf{F}_T^{3D}$ for voxel $P$ is computed as:
\begin{equation}
    \mathbf{F}^{3D}_T(P) = \mathrm{Deform3D}(\mathrm{RoI}(\mathbf{f}_T, \text{box}_P), D_p)
\end{equation}
Here, $\text{box}_P$ denotes the 2D bounding box corresponding to voxel centered on $P$ obtained via projection. This initialization yields a sparse but semantically filtered 3D representation, well-suited for downstream refinement and completion.

\subsection{Dummy Shortcut for Feature Refinement} \label{sec:dsfr}
\begin{figure*}[!htbp]
    \centering
    %\parbox{6.5in}
    {\includegraphics[width=.9\linewidth]{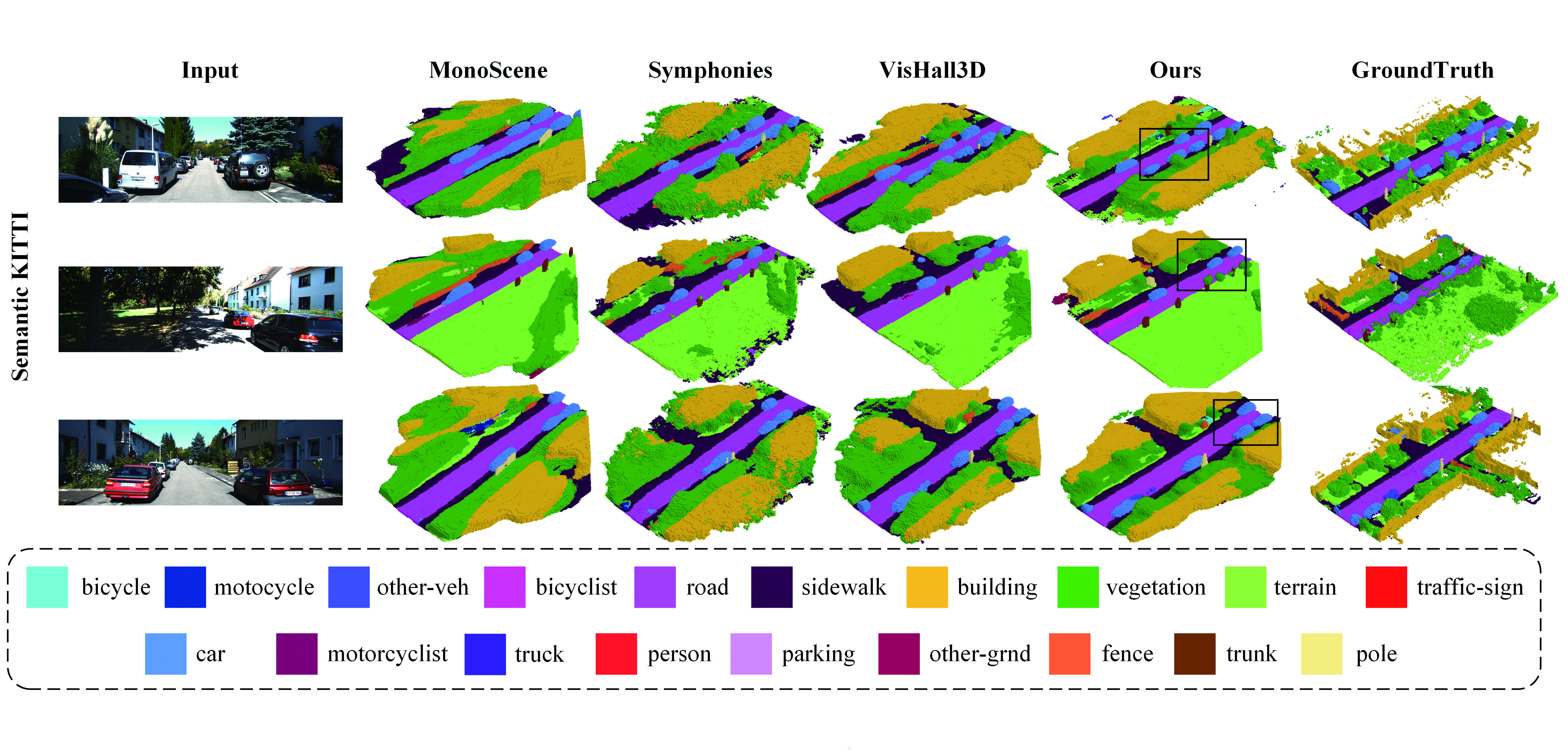}}
    \caption{Result of VoxSAMNet. Qualitative visual comparisons with Monoscene~\cite{cao2022monoscene}, Symphonies~\cite{jiang2024symphonize}, VisHall3D~\cite{Lu_2025_ICCV}, and Ground Truth (GT) on the SemanticKITTI~\cite{behley2019semantickitti} dataset.}
    \label{fig:Comparison}
\end{figure*}
After obtaining the initial 3D feature volume, we propose the \textbf{Dummy Shortcut for Feature Refinement (DSFR)} module to selectively refine voxel features based on their occupancy status. This design addresses the inherent sparsity of 3D scenes by avoiding redundant computation on empty voxels while enhancing the representation of informative regions.

The DSFR module consists of three main components: (i) voxel-wise occupancy classification, (ii) dummy feature assignment for empty voxels, and (iii) deformable attention-based refinement for occupied voxels.

\vspace{0.2em}\noindent\textbf{Voxel Classification.}\hspace{0.3em}
To distinguish between occupied and empty voxels, we apply a voxel-wise binary classifier that estimates the occupancy probability for each voxel. Given the initial 3D feature volume \( \mathbf{F}_{T}^{3D} \), we first extract spatial features via 3D convolutions, followed by a $1 \times 1\times 1$ convolution and sigmoid activation to obtain the occupancy probability map \( P_{\text{occu}} \in [0,1]^{H \times W \times D} \):
\begin{equation}
P_{\text{occu}} = \sigma(\text{Conv}_{1 \times 1\times 1}(\text{Conv}_{3D}(\mathbf{F}_{T}^{3D})))
\end{equation}

To classify voxels based on $P_{\text{occu}}$ , we introduce a learnable threshold parameter \(\theta\), which is mapped to a bounded range \([0.2, 1.0]\) via a sigmoid function: $\tau = 0.2 + 0.8 \cdot \sigma(\theta)$. Voxels with \( P_{\text{occu}} \geq \tau \) are classified as \emph{occupied}, while the rest are considered \emph{empty}. The corresponding binary masks are defined as:
\begin{equation}
M_o = \mathbb{I}(P_{\text{occu}}  \geq \tau), \quad M_e = \lnot M_o
\end{equation}
where $\mathbb{I}$ is the indicator function. This adaptive thresholding enables a trade-off between sparsity-aware acceleration and preservation of informative voxels.

\vspace{0.2em}\noindent\textbf{Empty Voxel Refinement via Dummy Embedding.}
For voxels classified as empty, we bypass expensive refinement operations and instead assign a learnable \textbf{dummy feature} \( \mathbf{Em} \). To provide a degree of continuity, we blend the original voxel features with the dummy feature based on the occupancy probability. Specifically, for a voxel centered at \(P\) in the empty voxel set, we use \(P_{\text{emp}}(P)\) to denote its empty probability, computed as the complement of \(P_{\text{occu}}(P)\). The refinement is then formulated as:
\begin{equation}
\mathbf{F}_T^{3D}(P) \gets \mathbf{F}_T^{3D}(P) \odot P_{\text{occu}}(P) + \mathbf{Em} \odot P_{\text{emp}}(P)
\end{equation}
Here, \(P\) denotes a voxel in the empty voxel set, and \( \odot \) denotes Hadamard product. This formulation ensures consistent representations for empty regions while reducing computational overhead.

\vspace{0.2em}\noindent\textbf{Occupied Voxel Refinement via Deformable Attention.}
For voxels classified as occupied and within the image's field of view \( M_o \cap \text{Mask}(u,v) \), we refine their features using multi-head deformable attention. This module selectively samples informative image features from the 2D feature map \( \mathbf{f}_T \), conditioned on each voxel’s 3D context.

For each occupied voxel with feature \( \mathbf{F}_T^{3D}(P) \),we predict the corresponding sampling offsets and attention weights using two separate multi-layer perceptrons (MLPs):
\begin{align}
\Delta &= \text{MLP}_{\text{offset}}(\mathbf{F}_T^{3D}(P)) \\
\text{Att} &= \text{softmax}(\text{MLP}_{\text{weight}}(\mathbf{F}_T^{3D}(P)))
\end{align}
% $\dagger$ denotes methods that use an external depth estimator pre-trained with stereo supervision. 
\begin{table*}[htbp]
    \fontsize{9pt}{10pt}\selectfont
    \setlength{\tabcolsep}{1.5pt}
    \renewcommand{\arraystretch}{1.2}
    \centering % 整个表格居中
    \caption{Quantitative results on the SemanticKITTI hidden test set. \textbf{Bold} denotes the best performance.} % 居中标题
    \label{tab:smkitti}
    \begin{tabular}{c|c|cc|ccccccccccccccccccc}
        \toprule
        \textbf{Method} & \textbf{Date} & \textbf{IoU$\uparrow$} & \textbf{mIoU$\uparrow$}
        & \cls{clsRoad}{road}
        & \cls{clsSidewalk}{sidewalk}
        & \cls{clsParking}{parking}
        & \cls{clsOtherGround}{other-grnd.}
        & \cls{clsBuilding}{building}
        & \cls{clsCar}{car}
        & \cls{clsTruck}{truck}
        & \cls{clsBicycle}{bicycle}
        & \cls{clsMotorcycle}{motorcycle}
        & \cls{clsOtherVeh}{other-veh.}
        & \cls{clsVegetation}{vegetation}
        & \cls{clsTrunk}{trunk}
        & \cls{clsTerrain}{terrain}
        & \cls{clsPerson}{person}
        & \cls{clsBicyclist}{bicyclist}
        & \cls{clsMotorcyclist}{motorcyclist}
        & \cls{clsFence}{fence}
        & \cls{clsPole}{pole}
        & \cls{clsTrafficSign}{traf.-sign} \\
        \midrule
        \multicolumn{22}{@{}l}{\quad\textbf{Stereo / multi-view methods}} \\ \hline
        ScanSSC~\cite{bae2025three} & CVPR2025 & 44.54 & 17.40 & 66.2 & 35.9 & 35.1 & 12.5 & 25.3 & 27.1 & 3.5 & 3.5 & 3.2 & 6.1 & 25.2 & 11.0 & 30.6 & 1.8 & 5.3 & 0.7 & 20.5 & 8.4 & 8.9 \\
        VLScene~\cite{wang2025vlscene} & AAAI2025 & 45.14 & 17.52 & 64.7 & 34.7 & 32.4 & 13.1 & 27.3 & 26.1 & 6.5 & 4.2 & 3.8 & 8.3 & 26.4 & 10.0 & 29.4 & 2.8 & 5.1 & 0.9 & 20.0 & 8.9 & 8.4 \\
        \hline
        \multicolumn{22}{@{}l}{\quad\textbf{Temporal methods}} \\
        \hline
        SOAP~\cite{lee2025soap} & CVPR2025 & 46.09 & 19.09 & 63.6 & 36.2 & 36.8 & 17.2 & 28.7 & 28.9 & 3.3 & 5.0 & 5.3 & 7.0 & 29.8 & 15.1 & 32.3 & 3.7 & 2.3 & 0.2 & 22.5 & 11.7 & 13.1 \\
        FlowScene~\cite{wang2025learning} & NIPS2025 & 45.20 & 17.70 & 64.1 & 35.0 & 33.7 & 13.0 & 27.7 & 26.4 & 10.0 & 4.2 & 3.1 & 7.0 & 26.3 & 10.0 & 30.2 & 3.1 & 5.1 & 1.1 & 22.2 & 8.9 & 9.1 \\
        \hline
        \multicolumn{22}{@{}l}{\quad\textbf{Single-frame methods}} \\
        \hline
        MonoScene~\cite{cao2022monoscene} & CVPR2023 & 34.16 & 11.08 & 54.7 & 27.1 & 24.8 & 5.7 & 14.4 & 18.8 & 3.3 & 0.5 & 0.7 & 4.4 & 14.9 & 2.4 & 19.5 & 1.0 & 1.4 & 0.4 & 11.1 & 3.3 & 2.1 \\
        %VoxFormer-S & CVPR2023 & 42.95 & 12.20 & 53.9 & 25.3 & 21.1 & 5.6 & 19.8 & 20.8 & 3.5 & 2.6 & 0.7 & 3.7 & 22.4 & 7.5 & 21.3 & 1.4 & 2.6 & 0.2 & 11.1 & 5.1 & 4.9 \\
        %TPVFormer & CVPR2023 & 34.25 & 11.26 & 55.1 & 27.2 & 27.4 & 6.5 & 14.8 & 19.2 & 3.7 & 1.0 & 0.5 & 2.3 & 13.9 & 2.6 & 20.4 & 1.1 & 2.4 & 0.3 & 11.0 & 2.9 & 1.5 \\
        %Symphonize & CVPR2024 & 42.19 & 15.04 & 58.4 & 29.3 & 26.9 & 11.7 & 24.7 & 23.6 & 3.2 & 3.6 & 2.6 & 5.6 & 24.2 & 10.0 & 23.1 & \textbf{3.2} & 1.9 & \textbf{2.0} & 16.1 & 7.7 & 8.0 \\
        H2GFormer-S~\cite{wang2024h2gformer} & AAAI2024 & 44.20 & 13.72 & 56.4 & 28.6 & 26.5 & 4.9 & 22.8 & 23.4 & 4.8 & 0.8 & 0.9 & 4.1 & 24.6 & 9.1 & 23.8 & 1.2 & 2.5 & 0.1 & 13.3 & 6.4 & 6.3 \\
        CGFormer~\cite{yu2024context} & NIPS2024 & 44.41 & 16.63 & 64.3 & 34.2 & 34.1 & 12.1 & 25.8 & 26.1 & 4.3 & 3.7 & 1.3 & 2.7 & 24.5 & 11.2 & 29.3 & 1.7 & 3.6 & 0.4 & 18.7 & 8.7 & 9.3 \\
        SkimbaDiff~\cite{liang2025skip} & AAAI2025 & 46.95 & 13.78 & 55.7 & 30.6 & 20.4 & 5.5 & 25.4 & 26.1 & 2.3 & 2.8 & 0.8 & 3.1 & 28.5 & 10.4 & 23.4 & 1.1 & 0.3 & 0.0 & 17.3 & 5.2 & 3.0 \\
        MixSSC~\cite{wang2025mixssc} & TCSVT2025 & 44.34 & 14.20 & 58.4 & 27.8 & 25.8 & 8.2 & 23.9 & 23.8 & 5.5 & 1.8 & 1.8 & 3.4 & 25.7 & 8.9 & 25.9 & 1.9 & 2.0 & 0.4 & 12.8 & 5.6 & 5.8 \\
        LOMA~\cite{cui2025loma} & AAAI2025 & 43.10 & 15.10 & 57.8 & 31.8 & 32.2 & 9.5 & 25.3 & 24.9 & 4.1 & 1.7 & 1.7 & 6.4 & 25.6 & 8.7 & 24.7 & 1.4 & 1.7 & 0.6 & 16.8 & 6.5 & 6.1 \\
        DISC~\cite{liu2025disentangling} & ICCV2025 & 45.32 & 17.35 & 63.1 & 34.7 & \textbf{34.6} & \textbf{12.6} & 26.6 & 26.7 & 5.5 & 4.0 & \textbf{4.7} & \textbf{8.1} & 26.5 & 10.3 & 29.3 & 2.8 & 2.5 & 1.1 & 19.3 & 8.4 & 8.7 \\
        VisHall3D~\cite{Lu_2025_ICCV} & ICCV2025 & 46.50 & 17.46 & 64.6 & 34.1 & 32.0 & 12.5 & 26.9 & 26.7 & \textbf{7.5} & 2.9 & 3.3 & 6.2 & 27.3 & 12.5 & 28.0 & 2.3 & 5.1 & \textbf{1.9} & \textbf{19.5} & 9.2 & 9.2\\
        \hline
         Ours & CVPR2026 & \textbf{47.88} & \textbf{18.19} & \textbf{65.8} & \textbf{35.5} & 33.9 & 12.3 & \textbf{28.2} & \textbf{27.0} & 7.2 & \textbf{4.4} & 3.9 & 7.7 & \textbf{28.7} & \textbf{13.4} & \textbf{30.3} & \textbf{3.1} & \textbf{5.3} & 0.5 & 19.1 & \textbf{9.9} & \textbf{10.5}\\
        \bottomrule
    \end{tabular}
\end{table*}
Sampling locations are computed by adding offsets \( \Delta \) to the 2D projection \( \widetilde{\mathbf{u}} \) of the voxel center, followed by normalization:
\begin{equation}
\mathbf{F}_s = \text{GridSample}(\mathbf{f}_T, \text{Norm}(\widetilde{\mathbf{u}} + \Delta))
\end{equation}

The sampled image features \( \mathbf{F}_s \) are aggregated using the learned attention weights $\text{Att}$ to get a new feature $G^{3D}_T(P)$. Finally, the aggregated features are fused with the original voxel features through concatenation and an MLP projection:
\begin{equation}
\mathbf{F}_T^{3D} \gets \text{MLP}(\text{Concat}([\mathbf{F}_T^{3D}, \mathbf{G}_T^{3D}]))
\end{equation}

\subsection{Training Losses} \label{sec:loss}
In the VoxSAMNet model, we adopt the scene-class affinity loss from MonoScene~\cite{cao2022monoscene} to enhance the model’s semantic and geometric understanding of 3D scenes. The semantic scene completion loss is shown as follows:
\begin{equation}
\mathcal{L}_{ssc} = \mathcal{L}_{sem} + \mathcal{L}_{geo} + \mathcal{L}_{ce} + \mathcal{L}_{depth}
\end{equation}

To further refine the model’s ability to distinguish between empty and occupied voxels, we introduce Occupied Loss tailored to voxel occupancy prediction. Drawing inspiration from MonoScene~\cite{cao2022monoscene}, it incorporates three binary cross-entropy (BCE) terms, \( \mathcal{L}_p \), \( \mathcal{L}_r \) and \( \mathcal{L}_s \) to optimize the precision, recall and specificity of the scene’s geometry.
\begin{equation}
\mathcal{L}_{occ} = \mathcal{L}_p + \mathcal{L}_r + \mathcal{L}_s
\end{equation}

The overall training loss function is formulated as follows:
\begin{equation}
\mathcal{L}_{total} = \mathcal{L}_{ssc} + \mathcal{L}_{occ}
\end{equation}
By combining these loss components, our training objective reinforces both text-guided semantic accuracy and spatially-aware volumetric reasoning, ensuring high-quality semantic scene completion.
\section{Experiment}
\begin{table*}[htbp]
\centering
\fontsize{9pt}{10pt}\selectfont
\setlength{\tabcolsep}{1pt}
\renewcommand{\arraystretch}{1.2}
\caption{Quantitative results on the SSCBench-KITTI-360 test set. \textbf{Bold} denotes the best performance.}
\label{tab:kitti360}
% 修改列定义为居中对齐
\begin{tabular}{c|c|cc|cccccccccccccccccc}
 
\toprule
\textbf{Method} & \textbf{Date} & \textbf{IoU$\uparrow$} & \textbf{mIoU$\uparrow$}
& \cls{clsCar}{car}
& \cls{clsBicycle}{bicycle}
& \cls{clsMotorcycle}{motorcycle}
& \cls{clsTruck}{truck}
& \cls{clsOtherVeh}{other-veh.}
& \cls{clsPerson}{person}
& \cls{clsRoad}{road}
& \cls{clsParking}{parking}
& \cls{clsSidewalk}{sidewalk}
& \cls{clsOtherGround}{other-ground.}
& \cls{clsBuilding}{building}
& \cls{clsFence}{fence}
& \cls{clsVegetation}{vegetation}
& \cls{clsTerrain}{terrain}
& \cls{clsPole}{pole}
& \cls{clsTrafficSign}{traf.-sign}
& \cls{clsOtherStruct}{other-struct.}
& \cls{clsOtherObj}{other-object.} \\
\midrule
% LiDAR-based methods（使用\multicolumn覆盖竖线）
%\multicolumn{21}{@{}l}{\quad\textbf{LiDAR-based methods}} \\ \hline
% & CVPR2017 & 53.58 & 16.95 & 32.0 & 0.0 & 0.2 & 10.3 & 0.0 & 0.2 & 65.7 & 17.3 & 41.2 & 3.2 & 44.4 & 6.8 & 43.7 & 28.9 & 0.8 & 0.8 & 8.7 & 0.7 \\
%LMSCNet& 3DV\phantom{R}2020& 47.35 & 13.65 & 20.9 & 0.0 & 0.0 & 0.3 & 0.6 & 0.0 & 63.0 & 13.5 & 33.5 & 0.2 & 43.7 & 0.3 & 40.0 & 26.8 & 0.0 & 0.0 & 3.6 & 0.0 \\
%\hline

% Camera-based methods（使用\multicolumn覆盖竖线）
\multicolumn{21}{@{}l}{\quad\textbf{Single-frame methods}} \\ \hline
MonoScene~\cite{cao2022monoscene} & CVPR2023 & 37.87  & 12.31 & 19.3 & 0.4 & 0.6 & 8.0 & 2.0 & 0.9 & 48.4 & 11.4 & 28.1 & 3.3 & 32.9 & 3.5 & 26.2 & 16.8 & 6.9 & 5.7 & 4.2 & 3.1\\
% OccFormer行（含19个有效数值）
OccFormer~\cite{zhang2023occformer} & ICCV2023 & 40.27 & 13.81 & 22.6 & 0.7 & 0.3 & 9.9 & 3.8 & 2.8 & 54.3 & 13.4 & 31.5 & 3.6 & 36.4 & 4.8 & 31.0 & 19.5 & 7.8 & 8.5 & 7.0 & 4.6 \\
% VoxFormer行（含18个有效数值，末尾补-占位）
VoxFormer~\cite{li2023voxformer} & CVPR2023 & 38.76 & 11.91 & 17.8 & 1.2 & 0.9 & 4.6 & 2.1 & 1.6 & 47.0 & 9.7 & 27.2 & 2.9 & 31.4 & 5.0 & 29.0 & 14.7 & 6.5 & 6.9 & 3.8 & 2.4 \\
Symphonies~\cite{jiang2024symphonize} & CVPR2024 & 44.12 & 18.58 & \textbf{30.0} & 1.9 & 5.9 & \textbf{25.1} & \textbf{12.1} & \textbf{8.2} & 54.9 & 13.8 & 32.8 & \textbf{6.9} & 35.1 & 8.6 & 38.3 & 11.5 & 14.0 & 9.6 & \textbf{14.4} & \textbf{11.3}\\
MixSSC~\cite{wang2025mixssc} & TCSVT2025 & 45.33 & 17.15 & 26.3 & 3.6 & 4.5 & 14.8 & 5.4 & 5.8 & 57.9 & 13.6 & 35.2 & 3.6 & 37.9 & 7.0 & 35.2 & 22.1 & 11.4 & 12.8 & 6.56 & 5.0\\
LOMA~\cite{cui2025loma} & AAAI2025 & 46.35 & 18.28 & 27.6 & 2.6 & 3.6 & 11.5 & 7.5 & 5.5 & 58.6 & 15.8 & 37.5 & 4.8 & 41.2 & 8.4 & 37.7 & 20.3 & 14.6 & 16.4 & 9.0 & 6.5\\
VLScene~\cite{wang2025vlscene} & AAAI2025 & 46.08 & 19.10 & 29.0 & \textbf{4.7} & \textbf{7.7} & 18.3 & 7.6 & 7.4 & 60.1 & 17.4 & 39.0 & 6.0 & 42.1 & \textbf{9.6} & 36.5 & 24.8 & \textbf{17.0} & 18.8 & 10.5 & 6.5\\
\hline
Ours& CVPR2026 & \textbf{47.22} & \textbf{20.21} & 29.2 & 3.3 & 4.3 & 19.6 & 7.7 & 7.0 & \textbf{61.9} & \textbf{18.9} & \textbf{39.9} & 5.2 & \textbf{42.2} & 7.6 & \textbf{38.7} & \textbf{24.9} & 16.5 & \textbf{20.6} & 9.0 & 7.2\\

\bottomrule
\end{tabular}
\end{table*}
To validate the effectiveness of our proposed VoxSAMNet in semantic scene completion, we conduct comprehensive experiments on two standard outdoor benchmarks: SemanticKITTI~\cite{behley2019semantickitti} and SSCBench-KITTI-360~\cite{liao2022kitti}. We compare our method against recent state-of-the-art approaches and perform ablation studies to analyze the contribution of each model component, particularly those related to text-guided supervision and sparsity-aware refinement.
\subsection{Experimental Setup}
\noindent\textbf{Datasets.}
SemanticKITTI~\cite{behley2019semantickitti} provides dense voxel-level semantic annotations derived from the KITTI Odometry dataset, with 10 training sequences (3,834 samples) and 20 semantic categories, at an image resolution of $1280 \times 384$. SSCBench-KITTI-360~\cite{liao2022kitti} is a more recent and challenging extension with long-range scenes and 19 semantic categories, offering 7 training sequences (8,487 samples) with an image resolution of $1408 \times 384$. In both datasets, the 3D scene is represented as a voxel grid of size $256 \times 256 \times 32$, covering a space of $51.2 \times 51.2 \times 6.4$m with a voxel resolution of 0.2m.

\noindent\textbf{Network Setup and Training.}
We adopt Swin Transformer~\cite{liu2021swin} as the 2D backbone, extracting multi-scale features at downsampling ratios of $1/8$, $1/16$, $1/32$, and $1/64$, with each scale having 256 channels. The number of image scales $N$ is set to 4.
The model is trained for 30 epochs using the AdamW optimizer~\cite{loshchilov2017decoupled}, with an initial learning rate of $10^{-4}$, reduced by a factor of 10 at epochs 20 and 25. Training is performed with a batch size of 4 on 4 NVIDIA RTX A100 GPUs. During training, text prompts are dynamically sampled based on the retained foreground categories, enabling the model to learn under diverse text-guided supervision. Depth prediction and text-guided filtering are jointly optimized with the voxel classification task.
\subsection{Comparative Results}
We compare VoxSAMNet against recent state-of-the-art methods on the public SemanticKITTI~\cite{behley2019semantickitti} and SSCBench-KITTI-360~\cite{liao2022kitti} benchmarks. For a fair comparison, all methods are evaluated under consistent settings following the official protocols of each benchmark. For the hidden test set of SemanticKITTI, predictions are submitted to the official evaluation server, since the corresponding ground-truth annotations are not publicly available. The quantitative results are reported in Tables~\ref{tab:smkitti} and~\ref{tab:kitti360}, where the compared methods are organized according to their inference-time input configurations, including single-frame methods~\cite{jiang2024symphonize, wang2024h2gformer, wang2025mixssc, zhang2025out, liang2025skip, zhang2023occformer, zhang2026pointcot}, temporal methods~\cite{lee2025soap, wang2025learning, zhang2026not}, and stereo/multi-view methods~\cite{bae2025three, wang2025vlscene, zhang2025ascot}.

On the SemanticKITTI dataset, VoxSAMNet achieves an mIoU of 18.19\%, surpassing all existing monocular SSC methods. Compared to the most recent monocular method DISC~\cite{liu2025disentangling}, our model improves mIoU by 0.84\% and voxel-wise IoU by 2.56\%. Notably, despite relying solely on single-frame RGB input, VoxSAMNet outperforms recent multi-view methods such as VLScene~\cite{wang2025vlscene} and ScanSSC~\cite{bae2025three} by 0.67\% and 0.79\% in mIoU, and by 2.74\% and 3.43\% in IoU, respectively. These results highlight the strong representation capability of our text-guided filtering and dummy-selective refinement mechanisms.

Qualitative results are shown in Figure~\ref{fig:Comparison}, where VoxSAMNet is compared to MonoScene~\cite{cao2022monoscene}, Symphonies~\cite{jiang2024symphonize} and VisHall3D~\cite{Lu_2025_ICCV} on SemanticKITTI. Our model produces more complete and semantically accurate reconstructions, especially in large structures such as roads, buildings, and vehicles. These improvements further validate the effectiveness of our approach in capturing global context and preserving fine-grained geometry in a sparse voxel setting.

\subsection{Ablations}
We conduct ablation studies on the SemanticKITTI~\cite{behley2019semantickitti} test set to systematically assess the contribution of each key component in VoxSAMNet. The analysis focuses on two aspects: (1) the impact of individual architectural modules, and (2) the choice of the image backbone in influencing performance. Together, these studies verify both the effectiveness of the proposed modules and the robustness of the resulting performance gains, showing that the improvements brought by VoxSAMNet remain consistent across different backbone choices.
% --- 请确保在您的 .tex 文件开头导入了这些宏包 ---
% \usepackage{booktabs} % 用于 \toprule, \midrule, \bottomrule
% \usepackage{amssymb}  % 用于 \checkmark
% \usepackage{float}    % (如果需要 [H] 定位)
% --------------------------------------------------

\begin{table}[htbp] % 采用您提供的 [htbp] 定位
\fontsize{9pt}{12pt}\selectfont % 严格按照您要求的 9pt 字体大小
\centering
\setlength{\tabcolsep}{3.5pt} % 设置一个标准的列间距，避免"打勾"列太挤
% 定义7列：方法(l), 三个组件(ccc), 三个指标(rrr)
\caption{Ablation study of the architectural components on the test set of SemanticKITTI.} % 采用您指定的标题
\label{tbl:arch_component} % 推荐使用一个唯一的标签
\begin{tabular}{c ccc ccc} 
\toprule
Method & TGIF & FD & DSFR & IoU(\%) & mIoU(\%) & Params(M) \\
\midrule
Baseline & & & & 45.38 & 15.84 & 116.30M \\
1 & \checkmark & & & 45.96 & 16.27 & 130.52M \\
2 & \checkmark & \checkmark & & 46.91 & 16.66 & 130.52M \\
3 & & & \checkmark & 46.65 & 17.42 & 129.48M \\
4 & \checkmark & & \checkmark & 47.12 & 17.86 & 143.58M \\
5 & \checkmark & \checkmark & \checkmark & \textbf{47.88} & \textbf{18.19} & 143.58M \\
\bottomrule
\end{tabular}
\end{table}
\noindent\textbf{Ablation on Architectural Components.}
To evaluate the effectiveness of each module, we conducted a series of ablation studies, progressively integrating the proposed components: the Text-Guided Image Filter (TGIF), Foreground Dropout (FD), and Dummy Shortcut for Feature Refinement (DSFR), with results shown in Table~\ref{tbl:arch_component}. The baseline model achieves 45.38\% IoU and 15.84\% mIoU, serving as a reference point. When TGIF is incorporated, using text prompts corresponding to the semantic categories actually present in each scene, the mIoU increases to 16.27\%, indicating that text-guided filtering effectively suppresses irrelevant image regions and enhances semantic focus. Upon further introducing FD, which randomly removes supervision signals of certain foreground instances, the text input in TGIF is correspondingly adjusted to align with the modified supervision. This configuration improves the mIoU to 16.66\%, confirming that targeted dropout mitigates overfitting and promotes more generalized feature learning. Finally, integrating DSFR leads to a significant boost in performance, with mIoU reaching 18.19\%. This highlights the importance of voxel-level sparsity modeling through selective refinement and dummy-based shortcuts. Experiments with other combinations of the modules consistently confirm the effectiveness of the proposed components.

% --- 请确保在您的 .tex 文件开头导入了这些宏包 ---
% --------------------------------------------------

\begin{table}[htbp] % 采用您提供的 [htbp] 定位
\fontsize{9pt}{12pt}\selectfont % 严格按照您要求的 9pt 字体大小
\centering
% 采用您模板中的列间距
\setlength{\tabcolsep}{3.5pt} 
\caption{Backbone ablation on the SemanticKITTI test set.}
\label{tb5:backbone_ablation} % 推荐使用一个唯一的标签
\begin{tabular}{c c c c} 
\toprule
Backbone & TGIF & IoU(\%) & mIoU(\%) \\
\midrule
ResNet50 & & 44.94 & 15.67 \\
ResNet50 & \checkmark & \textbf{45.75} & \textbf{15.98} \\
\midrule % 我添加了一个中间线来区分不同的backbone
Swin Transformer & & 45.38 & 15.84 \\
Swin Transformer & \checkmark & \textbf{45.96} & \textbf{16.27} \\
\bottomrule
\end{tabular}
% --- 使用您指定的英文标题 ---
\end{table}
\noindent\textbf{Ablation on Backbone.}
We conducted experiments using two different backbones: ResNet and Swin Transformer. Since the backbone is only related to the TGIF module, we limited the ablation study to the baseline and the addition of the TGIF module. As shown in the experimental results in Table~\ref{tb5:backbone_ablation}, it can be observed that the performance difference between the ResNet and Swin Transformer baselines is not significant. After incorporating the TGIF module, the model's test results improved in both mIoU and IoU, demonstrating the effectiveness of the TGIF module.
Simultaneously, we observe that the TGIF module implemented with Swin Transformer achieves a slight performance improvement, likely attributable to its inherent image-text alignment capabilities. Ultimately, we selected the pre-trained Swin Transformer as our image backbone to leverage its strengths in cross-modal representation.
\begin{table}[htbp]
    \fontsize{9pt}{12pt}\selectfont % 严格按照您要求的 9pt 字体大小
    \centering
    \caption{Computational cost of DSFR. Unit: FLOPs (G).}
    \label{tab6:dsfr_cost}
    \setlength{\tabcolsep}{3.8pt}
    \renewcommand{\arraystretch}{1.15}
    \begin{tabular}{cccc}
        \toprule
        Refine Strategy & Total & Deformable Attention & Ratio (\%) \\
        \midrule
        Dense & 30.96 & 24.512 & 79.2 \\
        Sparsity-Aware & \textbf{8.694} & \textbf{0.377} & \textbf{4.3} \\
        \midrule
        Reduction & \textbf{71.9\%} $\downarrow$ & \textbf{98.5\%} $\downarrow$ & - \\
        \bottomrule
    \end{tabular}
\end{table}

% \begin{table}[t]
%     \centering
%     \caption{Computational cost of DSFR. Unit: FLOPs (G).}
%     \label{tab6:dsfr_cost}
%     \setlength{\tabcolsep}{5pt}
%     \renewcommand{\arraystretch}{1.15}
%     \begin{tabular}{lcc}
%         \toprule
%         Refine Strategy & Total & DFA \\
%         \midrule
%         Dense & 30.96 & 24.512 \\
%         Sparsity-Aware & \textbf{8.694} & \textbf{0.377} \\
%         \midrule
%         Reduction             & \textbf{71.9\%} $\downarrow$ & \textbf{98.5\%} $\downarrow$ \\
%         \bottomrule
%     \end{tabular}
% \end{table}
\subsection{Computational Cost}
We evaluated the computational efficiency of VoxSAMNet on the SemanticKITTI validation set to assess its practical runtime performance. Our method achieves comparable computational efficiency to SOTA methods, with an inference time of 284ms per frame. Notably, it runs faster than most mainstream approaches, including Symphonize~\cite{jiang2024symphonize} (310ms) and OccFormer~\cite{zhang2023occformer} (338ms).

To validate the computational efficiency brought by the proposed DSFR module, we conduct a detailed FLOPs comparison as summarized in Table~\ref{tab6:dsfr_cost}. In the baseline configuration, where all voxels are processed through deformable attention, the total computational cost reaches 30.96 GFLOPs. Notably, deformable attention alone accounts for 24.512 GFLOPs, contributing a substantial 79.2\% of the overall computation, clearly identifying it as the primary bottleneck.

In contrast, the DSFR design fundamentally reshapes how computational resources are allocated. By applying deformable attention \emph{only} to occupied voxels—where informative geometric and semantic cues exist—the overall cost is dramatically reduced. This avoids spending expensive computation on the large number of empty voxels that contribute limited information to feature refinement. Under this setting, the total FLOPs drop to merely 8.694 GFLOPs. More impressively, the deformable attention computation falls to 0.377 GFLOPs, accounting for only 4.3\% of the total computation, yielding an overall reduction of 71.9\%.

This substantial improvement highlights the core advantage of our DSFR module: it introduces a fine-grained, content-aware computation strategy that selectively allocates attention capacity to informative regions while maintaining structural consistency for empty voxels with an extremely lightweight update rule. Such a design not only preserves the representational completeness of the voxel grid but also achieves a favorable cost--performance balance, making DSFR an efficient and scalable component for 3D perception pipelines.

\section{Conclusion}
We propose VoxSAMNet, an innovative monocular semantic scene completion framework that seamlessly integrates text-guided filtering with voxel-wise sparse refinement. By combining semantic priors from language with structured 3D voxel reasoning, the proposed framework produces more discriminative scene representations from a single RGB image. Specifically, the introduction of TGIF, foreground dropout, and DSFR modules collectively enables more effective semantic guidance, improved generalization, and efficient voxel processing. Experiments on both SemanticKITTI and SSCBench-KITTI-360 demonstrate that VoxSAMNet achieves state-of-the-art performance among monocular methods, confirming the effectiveness of the overall design. This work highlights the potential of language-guided 3D scene understanding and opens new avenues for advanced multimodal voxel reasoning.
%We propose VoxSAMNet, an innovative monocular semantic scene completion framework that seamlessly integrates text-guided filtering with voxel-wise sparse refinement. The introduction of TGIF, foreground dropout, and DSFR modules collectively enables more effective semantic guidance, improved generalization, and efficient voxel processing. Experiments on both SemanticKITTI and SSCBench-KITTI-360 demonstrate that VoxSAMNet achieves state-of-the-art performance among monocular methods. This work highlights the potential of language-guided 3D scene understanding and opens new avenues for advanced multimodal voxel reasoning.
% aaai原版如下
% We propose VoxSAMNet, a monocular semantic scene completion framework that integrates text-guided filtering with voxel-wise sparse refinement. The introduction of TGIF, foreground dropout, and DSFR modules enables effective semantic guidance, improved generalization, and efficient voxel processing. Experiments on SemanticKITTI and SSCBench-KITTI-360 demonstrate that VoxSAMNet achieves state-of-the-art performance among monocular methods. This work highlights the potential of language-guided 3D scene understanding and opens new avenues for advanced multimodal voxel reasoning.
% sec/6_ack.tex
\section*{Acknowledgements}
This paper is supported by the National Natural Science Foundation of China (No.62495084, No.62473306, No.U24B20181, No.62441616).
{
    \small
    \bibliographystyle{ieeenat_fullname}
    \bibliography{main}
}


\end{document}